\documentclass[letterpaper,10pt,journal,twoside]{IEEEtran}

\IEEEoverridecommandlockouts  

\usepackage{microtype}
\usepackage{amsmath, amssymb}
\usepackage{graphicx}
\usepackage{stfloats}
\usepackage[ruled,vlined,noend]{algorithm2e}
\usepackage{booktabs, multirow}
\usepackage{balance}
\usepackage[colorlinks]{hyperref}
\usepackage{color}
\usepackage{float}
\usepackage{gensymb}
\usepackage[usenames,dvipsnames]{xcolor}
\usepackage{enumitem}

\usepackage{pifont}
\newcommand{\cmark}{\ding{51}}
\newcommand{\xmark}{\ding{55}}

\title{DisMech: A Discrete Differential Geometry-based \\ Physical Simulator for Soft Robots and Structures}

\author{Andrew Choi$^{1}$, Ran Jing$^{2}$, Andrew Sabelhaus$^{2}$, and Mohammad Khalid Jawed$^{\dagger, 3}$

\thanks{Manuscript received November 16, 2023; accepted January 22, 2024. This letter was recommended for publication by Associate Editor R. Chandra and Editor A. Bera upon evaluation of reviewers' comments. This work was supported in part by the National Science Foundation under Grants OAC-2209782, OAC-220978, CAREER-2047663, and IIS-1925360.}
\thanks{$^{1}$Andrew Choi is with the Computer Science Department at the University of California, Los Angeles (UCLA) (email: {\tt \footnotesize asjchoi@cs.ucla.edu}).}%
\thanks{$^{2}$Ran Jing and Andrew Sabelhaus are with the Department of Mechanical Engineering at Boston University (email: {\tt \footnotesize \{rjing\}\{asabelha\}@bu.edu}).}%
\thanks{$^{3}$Mohammad Khalid Jawed is with the Department of Mechanical \& Aerospace Engineering at UCLA (email: {\tt \footnotesize khalidjm@seas.ucla.edu}).}%
\thanks{$^\dagger$ Corresponding author.}
\thanks{Digital Object Identifier 10.1109/LRA.2024.3365292}
}

\begin{document}

\urlstyle{tt}

\maketitle

\begin{abstract}
Fast, accurate, and generalizable simulations are a key enabler of modern advances in robot design and control. However, existing simulation frameworks in robotics either model rigid environments and mechanisms only, or if they include flexible or soft structures, suffer significantly in one or more of these performance areas. To close this ``sim2real'' gap, we introduce DisMech, a simulation environment that models highly dynamic motions of rod-like soft continuum robots and structures, quickly and accurately, with arbitrary connections between them. Our methodology combines a fully implicit discrete differential geometry-based physics solver with fast and accurate contact handling, all in an intuitive software interface. Crucially, we propose a gradient descent approach to easily map the motions of hardware robot prototypes to control inputs in DisMech. We validate DisMech through several highly-nuanced soft robot simulations while demonstrating an order of magnitude speed increase over previous state of the art. Our real2sim validation shows high physical accuracy versus hardware, even with complicated soft actuation mechanisms such as shape memory alloy wires. With its low computational cost, physical accuracy, and ease of use, DisMech can accelerate translation of sim-based control for both soft robotics and deformable object manipulation.

\end{abstract}


\section{Introduction}
\label{sec:introduction}

\IEEEPARstart{D}{eformable} materials are ubiquitous, from knots to clothing to new types of machines.
Despite the prevalence of deformable materials in everyday life, there is a lack of physically accurate and efficient continuum mechanics simulations for complicated and arbitrarily-shaped robots, particularly in hardware. 
As opposed to their rigid-body counterparts, deformable structures such as rods and shells possess infinite degrees of freedom and are capable of undergoing highly nonlinear geometric deformations from even minute external forces, requiring specialized simulators.

Accurate and efficient soft physics simulators serve two key purposes in the robotics community: they allow for 1) training traditional rigid manipulators to intelligently handle deformable objects~\cite{lin2021softgym} with sim2real realization~\cite{choi2024learning, tong2023sim2real} and 2) for the modelling, training, and development of controllers for soft robots themselves~\cite{huang2020rolling}. 
As the demands for efficient large-scale robot learning necessitate viable sim2real strategies, the need for such simulators becomes even more pressing.
With this in mind, we introduce DisMech, a full end-to-end discrete differential geometry (DDG)-based physical simulator for both soft continuum robots and structures.
Based on the Discrete Elastic Rods (DER)~\cite{bergou2008der, bergou2010dvt} framework, DisMech is designed to allow users to create custom geometric configurations composed of individual elastic rods.
Such configurations can be expressed quickly through an elegant API, allowing for rapid prototyping of different soft robot builds.
Actuation of the soft robot is then readily achievable by manipulating the natural curvatures of the individual elastic rods (i.e., limbs), enormously simplifying sim2real control tasks. 

As opposed to previous simulation frameworks focusing on soft physics~\cite{graule2022somogym, mathew2022sorosim, coevoet_2017_sofa_robot, gazzola2018forward}, DisMech's equations of motion are handled fully implicitly, allowing for far more aggressive time step sizes than previous formulations.
This allows our simulations to reach an order of magnitude speed increase over previous state-of-the-art simulators while maintaining physical accuracy as we show later for several canonical validation cases.
Though our prior work implemented a DDG-based simulation for soft robot locomotion~\cite{huang2020rolling}, this proof-of-concept operated solely in 2D, lacked elastic contact and self-contact, and was specialized to only one simple robot morphology.
In comparison, DisMech is a generalizable DDG-based framework for arbitrary soft rod-like robots in contact-rich 2D and 3D environments, and includes an algorithm to map hardware motions to intuitive DDG control inputs.
This article therefore compares a DDG simulation against the contemporary suite of soft robot simulators for the first time. 

This article's contributions are:
\begin{enumerate}
    \item We introduce a methodology and open-source implementation\footnote{\url{https://github.com/StructuresComp/dismech-rods}} of DisMech as a fully implicit simulator supporting soft physics, frictional contact, and intuitive control inputs. 
    To the best of our knowledge, DisMech is the first general purpose DDG-based 3D simulation framework for easy use by the robotics community.
    \item We numerically validate DisMech through several complicated simulations, and compare with the state-of-the-art framework Elastica~\cite{gazzola2018forward}, showing comparable accuracy while obtaining an order of magnitude speed increase.
    \item We demonstrate generalizability of DisMech for design simulations of dynamic soft robots, including a four-legged spider and an active entanglement gripper~\cite{Becker2022active_entanglement}. 
    \item Finally, we propose a generalizable gradient descent algorithm to map hardware robot data to DisMech's natural curvature parameters, validating their use as control inputs in ``real2sim'' of a real world soft manipulator~\cite{Pacheco2023comparison}.
\end{enumerate}

\section{Related Work}
\label{sec:related_work}

\begin{table*}
\renewcommand{\arraystretch}{1.2}
\centering
\caption{Robotics Simulation Framework Comparison}
\begin{tabular*}{\linewidth}{@{\extracolsep{\fill}} lccccc}
\toprule
\textbf{Framework} & Model & Continuum Physics & Physics Complexity &  Numerical Integration Scheme(s) & Solver Type \\
\midrule
Gazebo (ODE)~\cite{koenig2004gazebo}  & Rigid-body physics & \xmark & low & Semi-implicit Euler & Explicit \\
MuJoCo~\cite{emanuel_2012_mujoco}  & Rigid-body physics & \xmark & low & Semi-implicit Euler / 4$^\textrm{th}$ Runge-Kutta & Explicit or Implicit \\
Bullet~\cite{coumans2021_pybullet}  & Rigid-body physics  & \xmark & low & Semi-implicit Euler & Explicit \\
SoMo~\cite{graule2020somo}  & Rigid-body physics & \cmark${\boldsymbol{-}}$ & low & Semi-implicit Euler & Explicit \\
SOFA~\cite{Faure2012sofa, coevoet_2017_sofa_robot}  & FEA  & \cmark & high & Several Options & Explicit or Implicit \\
SoRoSim~\cite{mathew2022sorosim}  & Cosserat Rod + GVS  & \cmark & medium-high & Variable Step Methods  & Explicit or Implicit \\
Elastica~\cite{gazzola2018forward, zhang2019modeling}  & Cosserat Rod & \cmark & medium-high & Verlet Position & Explicit \\
DisMech  & Kirchhoff Rod + DDG & \cmark & medium-high & Backward Euler / Implicit Midpoint & Explicit or Implicit \\
\bottomrule
\end{tabular*}
\label{tab:comparison_frameworks}
\end{table*}

Though soft robotics research has been a popular and growing field~\cite{majidi_2014_soft_robots}, the presence of physical simulators capable of robustly capturing continuum mechanics is scarce. 
Popular simulation frameworks used for sim2real training of robots include Gazebo~\cite{koenig2004gazebo}, MuJoCo~\cite{emanuel_2012_mujoco}, and Bullet/PyBullet~\cite{coumans2021_pybullet}, but such simulators focus primarily on rigid body dynamics and fail to replicate physically accurate elasticity.

One avenue to remedy this involve frameworks such as SoMo~\cite{graule2020somo}, which models deformable structures by representing them as rigid links connected by springs.
Frameworks such as this simply act as wrappers built on top of preexisting rigid-body physics simulators (Bullet).
Although plausible results have been shown for soft grippers and control~\cite{graule2022somogym}, the underlying rigid-body physics is unable to simulate more complicated deformation modes such as twisting and elastic buckling~\cite{tong_2023_snap_buckling}. 
With this, accurate soft robot simulation requires simulation frameworks more dedicated to soft physics. 

Another avenue for simulating the continuum mechanics of soft robots involves the classical finite element method (FEM)~\cite{Faure2012sofa, coevoet_2017_sofa_robot}.
Although such methods are accurate, their high fidelity mesh representation often results in high computational costs, requiring separate model-order reduction (MOR) techniques for online control \cite{thieffry_control_2019,alora_data-driven_2023}.
These limitations focus the use of FEM/MOR for offline tasks such as reinforcement learning (RL) ~\cite{Schegg2022sofagym} and through asynchronous techniques~\cite{largilliere_2015_asynch_fem}.

Finally, there are simulation frameworks that simulate elastic rods using either the unstretchable and unshearable Kirchhoff rod model~\cite{kirchhoff1859uber} or fully deformable Cosserat rod model~\cite{gazzola2018forward}.
Using the Kirchhoff rod model, the computer graphics community has developed discrete differential geometry-based frameworks such as Discrete Elastic Rods (DER)~\cite{bergou2008der, bergou2010dvt}. 
Originally meant for realistic animation, DER has shown surprisingly great performance in accurately capturing the nonlinearity of rods and has been rigorously physically validated in cases such as rod deployment~\cite{jawed2014coiling}, knot tying~\cite{choi_imc_2021, tong_imc_2022}, and buckling~\cite{tong_2023_snap_buckling, tong2022helical_buckling}.
Using a DER-inspired framework, Huang et al. showed success at simulating and modeling soft robot locomotion~\cite{huang2020rolling}, albeit with the omission of twisting as locomotion occurred solely along a 2D plane.

Other frameworks have opted for using the Cosserat rod model as it incorporates the influence of shearing effects.
One such example involves SoRoSim~\cite{mathew2022sorosim} which models soft links using a Geometric Variable Strain (GVS) model of Cosserat rods.
This allows for the modeling of soft links with less degrees of freedom (DOFs) compared to traditional lumped mass models~\cite{bergou2008der, gazzola2018forward}.
Despite this mathematical elegance, the framework can be difficult for users to setup due to the prohibitively unintuitive representation of strains as the DOF.

Arguably the most prominent soft physics simulator based on Cosserat rod theory is Elastica~\cite{gazzola2018forward, zhang2019modeling}
This framework also models soft links as Cosserat rods but uses the more traditional lumped mass model similar to DER.
This framework has been extensively physically validated and its potential for reinforcement learning methods has also been demonstrated~\cite{Naughton2021_elastica_rl}.
Despite these results, Elastica uses an explicit integration scheme requiring prohibitively small time step sizes to maintain numerical stability.
This is especially true when simulating stiff materials as we will show later in Sec.~\ref{sec:validation}.

To circumvent this computational limitation, we formulate DisMech, a fully implicit DER-based physical simulator. 
Despite DER's non-consideration of shearing, we argue that shearing effects are negligible so long as the radii of the rod is small enough. 
Though this disqualifies DisMech from accurately modeling structures such as muscle fibers, the DDG-based framework of DisMech allows for an intuitive implicit formulation resulting in order of magnitude speed increases while maintaining highly accurate results for a large array of soft structures.
In addition to computational benefits, DisMech's framework also allows for the natural incorporation of shell structures via the DDG-based framework Discrete Shells~\cite{grinspun_2003_discrete_shells}, something that is rather nontrivial to accomplish in frameworks such as SoRoSim and Elastica.
A summary of popular robotics simulation frameworks can be seen in Table~\ref{tab:comparison_frameworks}.

\section{Methodology}
\label{sec:methodology}

\begin{figure}
    \centering
	\includegraphics[width=0.9\columnwidth]{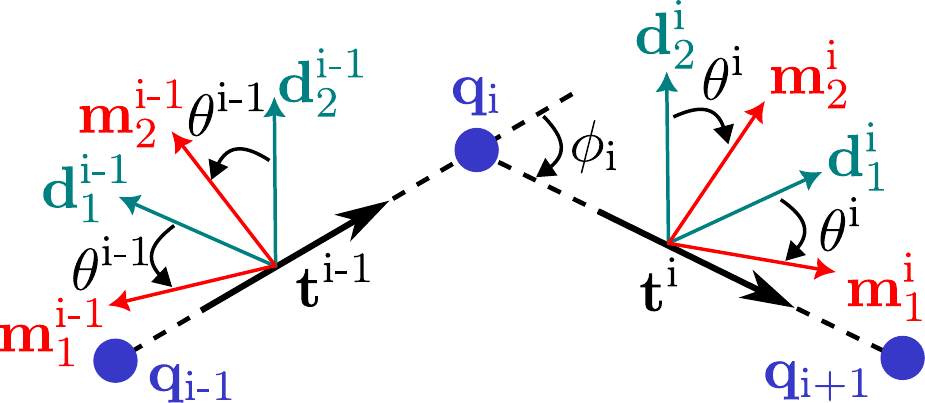}
	\caption{Discrete rod schematic with relevant notations describing the rod's discrete bending and twisting at node $\mathbf q_i$.} 
\label{fig:der_schematic}
\end{figure}

\begin{figure*}
	\includegraphics[width=\textwidth]{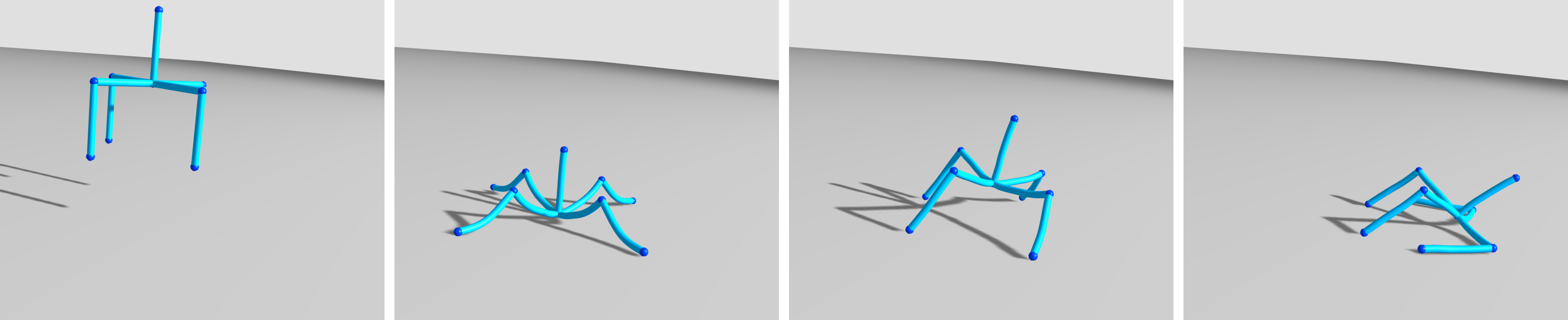}
	\caption{Rendering of a four-legged spider-like soft robot constructed using DisMech's API. Moving chronologically from left to right, the robot is dropped from a height where it then makes contact with an incline plane. After some initial sliding, the influence of sticking friction results in the robot's eventual equilibrium position in the rightmost column.} 
\label{fig:spider}
\end{figure*}

In this section, we formulate the DisMech framework involving its soft physics, frictional contact, and control input mapping.
As mentioned previously, DisMech's simulation of elasticity is based off of the Discrete Elastic Rods (DER)~\cite{bergou2008der, bergou2010dvt} framework.
In this framework, elastic rods are represented entirely by their centerline which are discretized into $N$ nodes $\mathbf q_i \in \mathbb R^3$, resulting in a total of $N-1$ edges $\mathbf e^i = \mathbf q_{i+1} - \mathbf q_{i}$.
Note that node-relevant quantities are denoted with subscripts while edge-relevant quantities are denoted with superscripts.
To capture the influence of bending and twist, each edge possesses two orthonormal frames: a reference frame $\{ \mathbf d_1^i, \mathbf d_2^i, \mathbf t^i \}$ and a material frame $\{ \mathbf m_1^i, \mathbf m_2^i, \mathbf t^i \}$ as shown in Fig.~\ref{fig:der_schematic}.
The material frame describes the rotation of the elastic rod centerline and is obtainable by rotating the reference frame by a signed angle $\theta^i$ along the shared director $\mathbf t^i$.
The reference frame is arbitrarily initialized at the start of the simulation and is updated between time steps via time parallel transport~\cite{bergou2010dvt}.
With $N$ Cartesian coordinate positions and $N-1$ twist angles, this results in a $4N-1$ size DOF vector
\begin{equation}
    \mathbf q = [\mathbf q_0, \theta^0, \mathbf q_1, \theta^1, ..., \mathbf q_{N-2}, \theta^{N-2}, \mathbf q_{N-1}]^T.
\end{equation}

\subsection{Elastic Energies}
\label{sec:elastic_energies}

We now define the constitutive laws of the elastic energies whose gradients and Hessians will be used to generate the inner elastic forces and Jacobians, respectively.
With DER based on the Kirchhoff rod model~\cite{kirchhoff1859uber}, the three main modes of deformation are stretching, bending, and twisting.

\subsubsection{Stretching Energy}

The stretching strain of an edge $\mathbf e^i$ is expressed simply by the ratio of elongation / compression with respect to its undeformed state.
Moving forwards, all quantities with a $\bar{}$ represent those respective quantities in their natural undeformed state.
With this, stretching strain can be defined as
\begin{equation}
    \epsilon^i = \frac{\lVert \mathbf e^i \rVert}{\lVert \bar{\mathbf e}^i \rVert} - 1.
\end{equation}

Stretching energy is then be computed as
\begin{equation}
    E_s = \frac{1}{2} EA \sum^{N-2}_{i=0} \left(\epsilon^i \right)^2 \lVert \bar{\mathbf e}^i \rVert,
\end{equation}
where $E$ is Young's modulus and $A$ is the cross-sectional area.

\subsubsection{Bending Energy}

Bending deformations occur between two adjacent edges and therefore, their respective strain involves computing the curvature at each interior node.
The bending strain can be evaluated through a curvature binormal which captures the misalignment between two edges:
\begin{equation}
    (\kappa \mathbf b)_i = \frac{2 \mathbf t^{i-1} \times \mathbf t^i}{1 + \mathbf t^{i-1} \cdot \mathbf t^i }.
\end{equation}

Using this curvature binormal, the integrated curvature vector at node $\mathbf q_i$ can be evaluated using the material frames of each adjacent edge as
\begin{equation}
\label{eq:compute_kappa}
\begin{split}
    \kappa_{1,i} = &\frac{1}{2} (\kappa \mathbf b)_i \cdot \left(\mathbf m^{i-1}_2 + \mathbf m^i_2 \right), \\
    \kappa_{2,i} = -&\frac{1}{2} (\kappa \mathbf b)_i \cdot \left(\mathbf m^{i-1}_1 + \mathbf m^i_1 \right), \\
    \boldsymbol \kappa_i = &\left[\kappa_{1, i}, \kappa_{2, i} \right]
\end{split}
\end{equation}

Note that $\boldsymbol \kappa_i   = 2 \tan (\boldsymbol \phi_i / 2)$ where $\boldsymbol \phi_i = [\phi_{1, i}, \phi_{2, i}]$ is the turning angle vector (Fig.~\ref{fig:der_schematic}).
With this, we compute the bending energy of the rod as
\begin{equation}
\label{eq:bending_energy}
    E_b = \frac{1}{2} EI \sum^{N-2}_{i=1} (\boldsymbol \kappa_i - \bar{\boldsymbol \kappa}_i)^2 \frac{1}{V_i},
\end{equation}
where $I = \pi h^4 / 4$ is the moment of inertia; $h$ is the rod radius, and $V_i = (\lVert \mathbf e^i \rVert + \lVert \mathbf e^{i-1} \rVert) / 2$ is the Voronoi length.
Later, we will show how to manipulate the natural curvature $\bar{\boldsymbol \kappa}$ to intuitively actuate the robot's limbs.

\subsubsection{Twisting Energy}

Like bending, twisting is also a deformation mode that occurs between two adjacent edges.
Therefore the twisting strain at an interior node $\mathbf q_i$ is computed as 
\begin{equation}
    \tau_i = \theta^i - \theta^{i-1} + \beta^i,
\end{equation}
where $\beta^i$ is the signed angular difference between the two consecutive references frames of the $i$ and $i-1$-th edges.

Finally, twisting energy is then defined as
\begin{equation}
    E_t = \frac{1}{2} GJ \sum^{N-2}_{i=1} \left( \tau_i - \bar \tau_i \right)^2 \frac{1}{V_i},
\end{equation} 
where $G$ is the shear modulus and $J = \pi h^4 / 2$ is the polar second moment of area.

\subsubsection{Elastic Forces}

With each of the elastic energies defined, we can now compute the inner elastic forces (for nodal positions $\mathbf q_i$) and elastic moments (for twist angles $\theta^i$) as the negative gradient of elastic energy:
\begin{equation}
    \mathbf F^\textrm{int} = - \frac{\partial}{\partial \mathbf q} (E_s + E_b + E_t).
\end{equation}

\subsection{Contact and Friction}

In addition to the inherent elasticity of the rod, we must also model external forces such as contact and friction.
We do so by using Implicit Contact Model (IMC)~\cite{choi_imc_2021, tong_imc_2022}, an implicit penalty energy method that integrates seamlessly into the DER framework.
Similar to the elastic energies, IMC defines an artificial contact energy 
\begin{equation}
\label{eq:contact_energy}
    E_c = \begin{cases}
                                (C - D)^2 & D \in (0, C-\delta], \\
                                \left(\frac{1}{K_1}\log\left(1 + e^{K_1 (C - D)}\right) \right)^2  & D \in (C - \delta, C + \delta), \\
                                0 &  D \geq C + \delta,
                                 \end{cases}
\end{equation}
where $K_1 = 15 / \delta$ is a stiffness parameter; $\delta$ is a user-defined contact distance tolerance; $D$ is the minimum distance between two edges, and $C$ is the distance from the centerlines at which contact would occur (e.g., self-contact occurs at $C=2h$).
By employing chain rule, we can obtain contact forces promoting non-penetration as 
\begin{equation}
    \mathbf F^\textrm{con} = -\dfrac{\partial E_c}{\partial D} \dfrac{\partial D}{\partial \mathbf q}.
\end{equation}

Coulomb friction forces for an edge $i$ are then computed using the above contact forces (i.e., the normal forces), as 
\begin{align}
    \gamma &= \dfrac{2}{1 + e^{-K_2 \lVert \mathbf u \rVert}} - 1, \\
    \mathbf F^{\textrm{fr}, i} &= - \mu \gamma \hat{\mathbf u} \lVert \mathbf F^{\textrm{con}, i} \rVert, 
\end{align}
where $K_2 = 15 / \nu$ is a stiffness parameter; $\nu$ is a user-defined slipping tolerance; $\mathbf u$ is the tangential relative velocity between two contacting bodies (note that $\hat{}$ indicates unit vector); $\mu$ is the friction coefficient, and $\gamma \in [0, 1]$ is a smooth scaling factor that models the transition from sticking and sliding modes between the contacting bodies.
This results in an external force vector of $\mathbf F^\textrm{ext} = \mathbf F^\textrm{con} + \mathbf F^\textrm{fr} + \mathbf F^\textrm{misc}$,
where $\mathbf F^\textrm{misc}$ refers to other miscellaneous external forces such as gravity, viscous damping, hydrodynamic forces, etc.

\subsection{Time Stepping and Numerical Integration Schemes}

With the internal and external forces defined, we can now construct the equations of motion (EOM) using Newton's second law as
\begin{equation}
\label{eq:simple_eom}
    \mathbf M \ddot{\mathbf q} = \mathbf F^\textrm{int} + \mathbf F^\textrm{ext},
\end{equation}
where $\mathbf M$ is the diagonal mass matrix and $\ddot{\mathbf q}$ is the second derivative of the DOFs with respect to time.
Given this, we can then format the EOM to be solved for implicitly as
\begin{align}
\label{eq:implicit_eom}
    &\frac{\mathbf M}{\Delta t} \left( \frac{\mathbf q(t_{i+1}) - \mathbf q(t_i)}{\Delta t} - \dot{\mathbf q}(t_i)  \right) - \mathbf F^\textrm{int} - \mathbf F^\textrm{ext} = 0, 
\end{align}
where $\Delta t$ is the time step size.
We solve for $\mathbf q(t_{i+1})$ in Eq.~\ref{eq:implicit_eom} iteratively using Newton's method using the elastic energy Hessian, contact energy Hessian, and friction Jacobian (obtainable via chain rule~\cite{tong_imc_2022}):
\begin{equation}
    \mathbf J = \dfrac{\partial^2}{\partial \mathbf q^2} (E_s + E_b + E_t) + \dfrac{\partial^2 E_c}{\partial \mathbf q^2} + \dfrac{\partial \mathbf F^\textrm{fr}}{\partial \mathbf q}.
\end{equation}
Once $\mathbf q(t_{i+1})$ is known, the velocity at time $t+1$ can be trivially solved for algebraically as $\dot{\mathbf q}(t_{i+1}) = (\mathbf q(t_{i+1}) - \mathbf q(t_i)) / \Delta t$.

Currently, DisMech has been outfitted to support both explicit (e.g., Elastica's Verlet position~\cite{gazzola2018forward}) and implicit numerical integration schemes.
For implicit schemes, we offer both backward Euler and implicit midpoint.
As backward Euler results in artificial damping~\cite{huang_2019_newmark}, we opt to use implicit midpoint for scenarios where energy loss is unwanted and  backward Euler for scenarios where stability is preferred. 
To further maintain numerical stability, DisMech is outfitted with both a line search method~\cite{tong_imc_2022} as well as adaptive time stepping as optional features when an implicit scheme is chosen.

\subsection{Elastic Joints}

To allow for creating connections and grid-like structures, we introduce the concept of ``elastic joints", i.e., nodes along a rod that have one or more other rods connected to them. Such connections experience the same stretching, bending, and twisting energy constraints formulated in Sec.~\ref{sec:elastic_energies} as individual rods. Bending and twisting forces at joints are computed for every possible edge combination.
Examples of elastic joints can be seen in Figs.~\ref{fig:spider} and~\ref{fig:real2sim} as blue spheres.

\subsection{Actuation via Natural Curvatures}
\label{sec:natural_curvature_control}
Given how the material frames are first setup, we can actuate a soft robot by manipulating its natural curvature $\bar{\boldsymbol \kappa}$ or equivalently its natural turning angle $\bar{\boldsymbol \phi}$. 
This in effect results in a bending energy differential in Eq.~\ref{eq:bending_energy} which produces contractions and/or relaxations of the elastic rod along the respective material directors.
Later in Sec.~\ref{sec:real2sim}, we show how this actuation can be framed as a control input, using a gradient descent method to calculate $\bar {\boldsymbol \kappa}$ values for real2sim open-loop control of a hardware robot prototype.

\section{Theoretical Validation}
\label{sec:validation}

\begin{table*}
\renewcommand{\arraystretch}{1.2}
\centering
\caption{Simulation Performance Comparison}
\begin{tabular*}{\linewidth}{@{\extracolsep{\fill}} l c c c ccccc }
\toprule
\multirow{5}{*}[-3pt]{\rotatebox[origin=c]{0}{\begin{tabular}{@{}c@{}}Simulation \\[-2pt] Framework\end{tabular}}} & \multirow{5}{*}[-3pt]{\begin{tabular}{@{}c@{}}Numerical \\[-2pt] Integration\end{tabular}} & \multirow{5}{*}[-3pt]{Metrics [s]} & \multicolumn{6}{c}{Sim Validation Experiments} \\
\cmidrule{4-9}
& & & Parameters & Cantilever (a1)  & Cantilever (a2) & Helix (b1) & Helix (b2)  & Friction (c) \\
& & & $N$ & $201$ & $201$ & $100$ & $100 $& $26$ \\
& & & $h$ [m] & $0.020$ & $0.020$ & $0.005$ & $0.001$ & $0.025$ \\
& & & $E$ [Pa] & $1 \times 10^5$& $1 \times 10^7$ & $1 \times 10^7$ & $1 \times 10^9$ & $1 \times 10^5$ \\
& & & $T$ [s] & $100$ & $20$ & $10$ & $2$ & $1.5$ (44 sims) \\
\midrule
\multirow{2}{*}[-1pt]{Elastica}  &   \multirow{2}{*}[-0pt]{\begin{tabular}{@{}c@{}}Verlet \\[-0pt] Position\end{tabular}}   &  Max $\Delta t$ & | &  $2.5\times 10^{-4}$ & $3 \times 10^{-5}$ & $4\times10^{-5}$& $1\times 10^{-6}$& $1\times 10^{-4}$ \\
& & Comp. Time & | & $23.80$ & $40.01$ & $11.77$ & $89.24$& $7.42$ \\
\multirow{2}{*}[-1pt]{DisMech} & \multirow{2}{*}[-0pt]{\begin{tabular}{@{}c@{}}Implicit \\[-0pt] Midpoint\end{tabular}} & Best $\Delta t$ & | & $5 \times 10^{-1}$& $5 \times 10^{-2}$& $6 \times 10^{-3}$ & $3\times10^{-3}$& $5\times 10^{-3}$\\
& & Comp. Time & | & $0.661$ & $1.24$& $7.30$ & $3.18 $& $1.59$  \\
\bottomrule
\end{tabular*}
\label{tab:performance_comparison}
\end{table*}

\begin{figure*}
	\includegraphics[width=\textwidth]{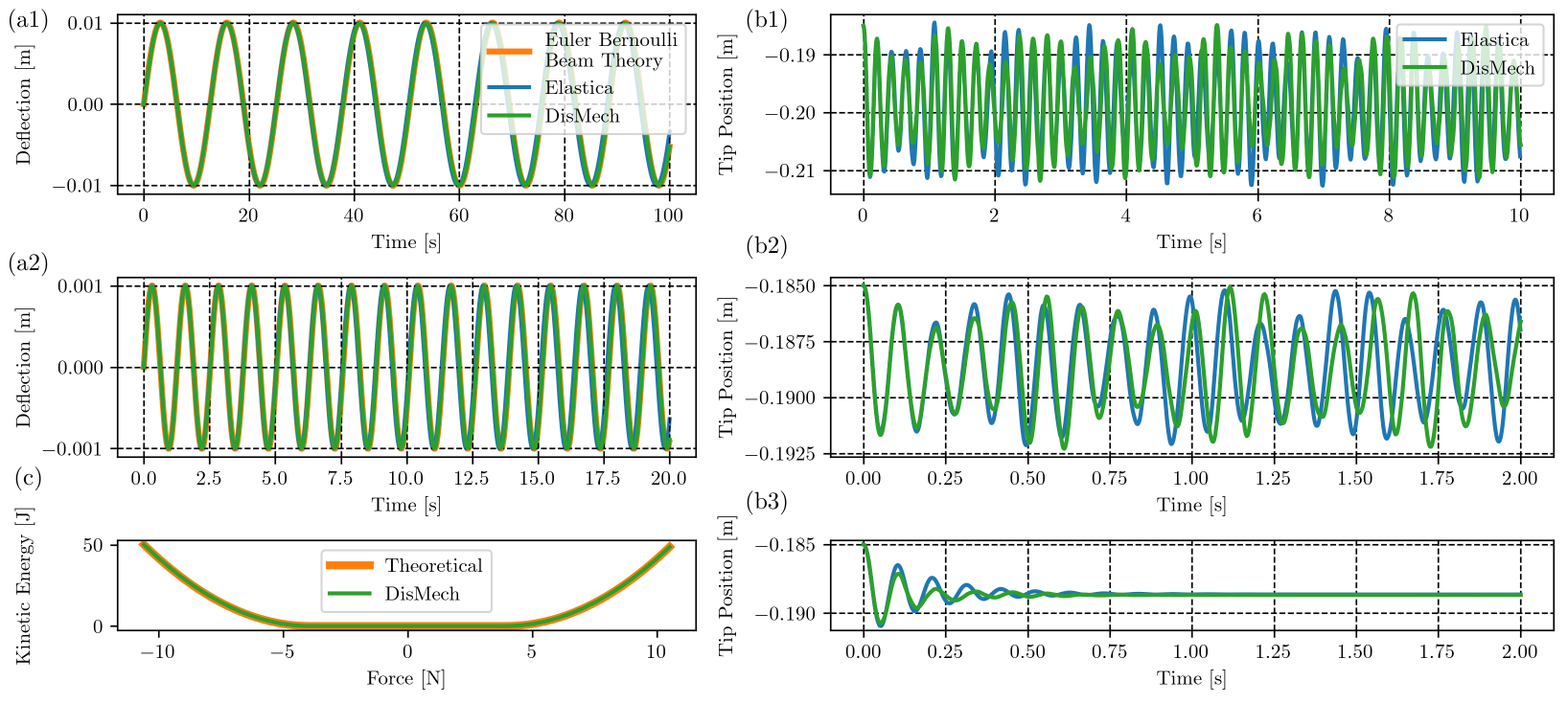}
	\caption{Simulation results for DisMech compared with Elastica and theory (when applicable). 
Each plot refers to a specific experiment shown in Table~\ref{tab:performance_comparison}. Plots (a1) and (a2) showcase the deflection of the cantilever beam experiment for different material properties. 
 Likewise, plots (b1) and (b2) showcase the tip position of the helical rod experiment for different material and geometric properties.
 As the helix experiment does not have an analytical solution to compare against, we show in plot (b3) that both DisMech and Elastica reach the same static equilibrium point when damping is introduced into the system as a sanity check.
 Finally, plot (c) showcases the kinetic energy of the rod as a function of an enacted external force for a friction coefficient of $\mu = 0.4$.
  } 
\label{fig:results}
\end{figure*}

In this section, we perform physical validation of our model through several canonical experiments.
We compare our results with both theory (when appropriate) and the state-of-the-art framework Elastica~\cite{gazzola2018forward}.
Efficiency metrics such as time step size and computational time for all experiments are listed in Table~\ref{tab:performance_comparison} while simulation results are plotted in Fig.~\ref{fig:results}.
For Elastica, the largest time step size before numerical instability arose was used while for DisMech, the time step size that had the best tradeoff between number of iterations and computational time was used.
For all experiments, we assume a Poisson ratio of $0.5$ and a gravitational acceleration of $g=9.8$m/s$^2$ for those involving gravity.
All simulations were run on a workstation containing an Intel Core i7-9700K (3.60GHz$\times$8) processor and 32GB of RAM.

\subsection{Dynamic Cantilever Beam}

The first validation experiment we conduct is comparing the deflections of a rod modelled as a cantilever beam against Euler-Bernoulli beam theory. 
In absence of external loads, we can refer to the free vibration equation
\begin{equation}
\label{eq:dynamic_beam}
    w(s, t) = \textrm{Re} \left[\hat w(s) e^{-i \omega t} \right],
\end{equation}
where $w$ is the $z$-direction deflection at arc length $s \in [0, L]$ at time $t$; $\omega$ is the frequency of the vibration, and $\hat w(x)$ is the natural frequency of the beam.
We use the analytical solution available for cantilever beams~\cite{han1999_dynamic_beam} for an initial tip velocity of $5$\,mm/s.
With this, we conduct two simulations to showcase the effect of material stiffness on time step size.
For both experiments, we use a rod radius $h=2$\,cm, density $\rho=500$\,kg/m$^3$, and rod length $L=1$\,m.
A discretization of $N=201$ nodes is also used.
For the first and second experiments, we then use a Young's modulus $E$ of $0.1$\,MPa and $10$\,MPa and a sim time $T$ of $100$\,s and $20$\,s, respectively.

Deflections in Fig.~\ref{fig:results}(a1-a2) show that both Elastica and DisMech have excellent agreement with theory (Eq.~\ref{eq:dynamic_beam}), and neither suffer from artificial damping.
However, DisMech achieves these results with time step sizes three orders-of-magnitude larger than Elastica, corresponding to an order of magnitude speed increase.

\begin{figure*}
	\includegraphics[width=\textwidth]{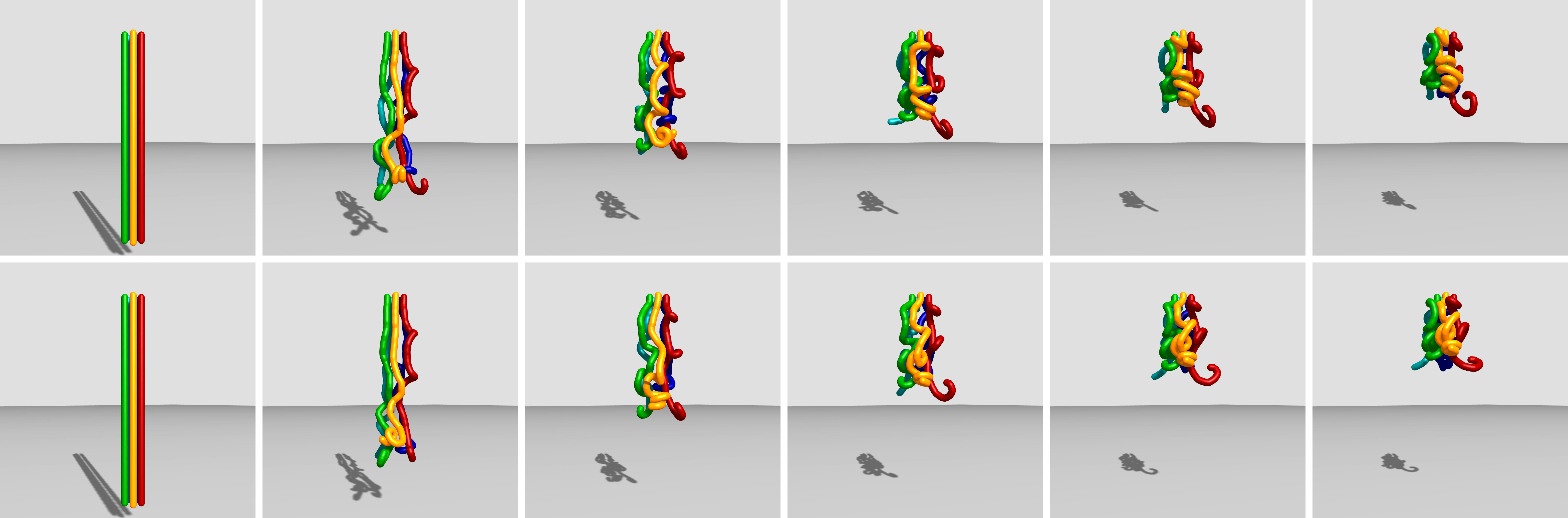}
	\caption{Renderings of an active entanglement gripper~\cite{Becker2022active_entanglement}. The top row showcases contact-only entanglement whereas the bottom row showcases entanglement with friction $\mu=0.5$. Note the influence of stiction causing more distorted helices in the latter example.}
\label{fig:active_entanglement}
\end{figure*}

\subsection{Oscillating Helix under Gravity}

The next experiment consists of a suspended helical rod oscillating under gravity.
We use the same experiment setup as~\cite{huang_2019_newmark} with density $\rho=1273.52$ \,kg/m$^3$, helix radius of $2$\,cm, pitch of $5$\,cm, and a contour length of $0.5$\,m (resulting in an axial length of $\approx 0.185$\,m).
A discretization of $N=100$ is used.
Like the previous section, we also conduct two experiments with a Young's modulus $E$ of $10$\,MPa and $1$\,GPa, radius $h$ of $5$\,mm and $1$\,mm, and sim time $T$ of $10$\,s and $2$\,s, respectively.

As an analytical solution for comparison is not available, we simply plot the oscillating tip positions as shown in Fig.~\ref{fig:results}(b1-b2).
As shown, both simulation frameworks produce results with identical frequencies for the first experiment and near identical frequencies for the second.
To confirm that the difference in results is not a result of improper system representation, we also show that both DisMech and Elastica reach the same static equilibrium point in Fig.~\ref{fig:results}(b3) after introducing damping.
Therefore, we presume that these slight frequency differences are numerical artifacts arising from differences in integration schemes and/or time step size.
As before, we also report being able to take time step sizes of up to three orders-of-magnitude, correlating to an order of magnitude speed increase.

\subsection{Friction Validation}

For the final experiment, we test axial friction on a rod.
We use a rod with radius $h=2.5$\,cm, density $509.3$\,kg/m$^3$, length $L=1$\,m, and Young's modulus $E=0.1$\,MPa. A discretization of $N=26$ is used. Floor contact was simulated using IMC using a friction coefficient $\mu=0.4$, distance tolerance $\delta=5\textrm{e}{-4}$\,m, and slipping tolerance $\nu=1\textrm{e}{-3}$\,m/s.
With this, we can compute the kinetic energy of the rod when experiencing a constant uniform push/pull force $\mathbf F^\textrm{p}$ as
\begin{equation}
    E_{k} = \begin{cases}
        0 & \lVert \mathbf F^\textrm{p} \rVert \leq | \mu m g |, \\
        \dfrac{t^2}{2m} \left(\lVert \mathbf F^\textrm{p} \rVert - \mu m g  \right)^2 & \lVert \mathbf F^\textrm{p} \rVert > | \mu m g |,
    \end{cases}
\end{equation}
where $m$ is the mass of the rod and $t$ is the time after starting force exertion from a resting configuration.
We compute results for 44 simulations in parallel for a range of forces $\lVert \mathbf F^\textrm{p} \rVert \in [-10.6, 10.5]$\,N.
When observing results in Fig.~\ref{fig:results}(c), we see excellent agreement between DisMech's simulated kinetic energy and theory for a sim time of $T=t=1.5$\,s.

\section{Practical Demonstrations for Flexible Robots}
\label{sec:use_cases}

Three demonstrations below showcase the accuracy and ease-of-use of DisMech over competing frameworks in simulating complicated flexible and soft robots.
We also provide an algorithm that maps hardware robot motions to DisMech's control inputs, emphasizing its generalizability.

\subsection{Arbitrary Robot Prototyping}
\subsubsection{Spider Robot}
We showcase the ease-of-use and environmental contact capabilities of DisMech by simulating a four-legged spider-like soft robot composed of interconnected rods (Fig.~\ref{fig:spider}).
Individual limbs are created as rods, with joints between them, using a single line of code each. 
We simulate dropping the robot onto a floor (no actuation) with $\mu=0.4$, and replicate an incline using gravity $\mathbf g = [0.707, 0.707, -9.8]$\,m/s$^2$.
We can observe visually plausible results of the robot colliding with the floor, rebounding, and then after an initial sliding period, coming to static equilibrium via stiction.
\subsubsection{Active Entanglement Gripper}
Next, we further showcase the generality of DisMech by simulating an active entanglement gripper~\cite{Becker2022active_entanglement}, a highly nontrivial frictional contact case.
To do so, we simulate five fingers (rods) placed equidistantly along a circular perimeter, each with length $L=0.3$\,m, radius $h=5$\,mm, density $\rho=1200$\,kg/m$^3$, and Young's modulus $E=0.3$\,MPa.
Each rod is discretized using $N=60$.
Contact is simulated using $\delta=5\textrm{e}{-4}$\,m and $\nu=1\textrm{e}{-3}$\,m/s.
To simulate rapid entanglement, each edge is actuated via a random $\bar \phi$ value uniformly sampled from a range of $[0, 40]\,\degree$.

Results for both contact-only and frictional contact ($\mu=0.5$) scenarios show convincing, visually plausible results (Fig.~\ref{fig:active_entanglement}). 
When comparing between the two, friction causes the rods to stick to each other during the coiling, resulting in more distorted helices.
Simulations such as these open up many opportunities for generating control policies.

\begin{algorithm}
\SetAlgoLined
\LinesNumbered
\DontPrintSemicolon
\KwIn{$\mathbf D \gets$ geometric configuration of robot}
\KwOut{${\boldsymbol \tau}_{\bar{\boldsymbol \kappa}} \gets$ trajectory of $\bar{\boldsymbol \kappa}$s}
\SetKwFunction{ComputeCurvature}{ComputeCurvature}
\SetKwProg{Fn}{Func}{:}{}
\SetKwFunction{SolveCurvatures}{SolveNaturalCurvatures}
{
\Fn{\SolveCurvatures$(\mathbf D)$}
{
$\boldsymbol{\kappa}^* \gets \ComputeCurvature(\mathbf D)$ \tcp*{Eq.~\ref{eq:compute_kappa}}
${\boldsymbol \tau}_{\bar{\boldsymbol \kappa}} \gets [ \ \ ]$ \tcp*{trajectory}

$\bar{\boldsymbol \kappa} \gets [0, 0]$\;
\For{$\kappa_1^*, \kappa_2^* \in \boldsymbol{\kappa}^*$}{

$\alpha \gets 0.1$ \tcp*{step size}
$\lambda_\textrm{prev} \gets \infty$ \;

\While{\textup{True}}{
    $\kappa_1, \kappa_2 \gets \mathcal F(\bar{\boldsymbol \kappa})$ \tcp*{DisMech Sim}
    $\lambda \gets | \kappa_1^* - \kappa_1 | + | \kappa_2^* - \kappa_2 | $ \tcp*{Eq.~\ref{eq:lambda}}

    \If{$\lambda < \textup{tolerance}$}{
        break \;
    }

    \If{$\lambda > \lambda_\textup{prev}$}{
        $\alpha \gets 0.5\alpha$
    }
    $\lambda_\textrm{prev} \gets \lambda$ \;

    $\nabla_{\boldsymbol \kappa} \lambda \gets $ finite diff with $\epsilon$ \tcp*{Eq.~\ref{eq:finite_diff}}

    $\bar{\boldsymbol \kappa} \gets \bar{\boldsymbol \kappa} - \alpha \nabla_{\boldsymbol \kappa} \lambda$ \tcp*{Eq.~\ref{eq:grad_descent_update}}
    
}
${\boldsymbol \tau}_{\bar{\boldsymbol \kappa}}$.append$\left(  \bar{\boldsymbol \kappa}  \right)$ \;

}
\textbf{return} ${\boldsymbol \tau}_{\bar{\boldsymbol \kappa}}$ \;
}
}
\caption{Real2Sim via Gradient Descent}
\label{alg:real2sim}
\end{algorithm}

\begin{figure*}
	\includegraphics[width=\textwidth]{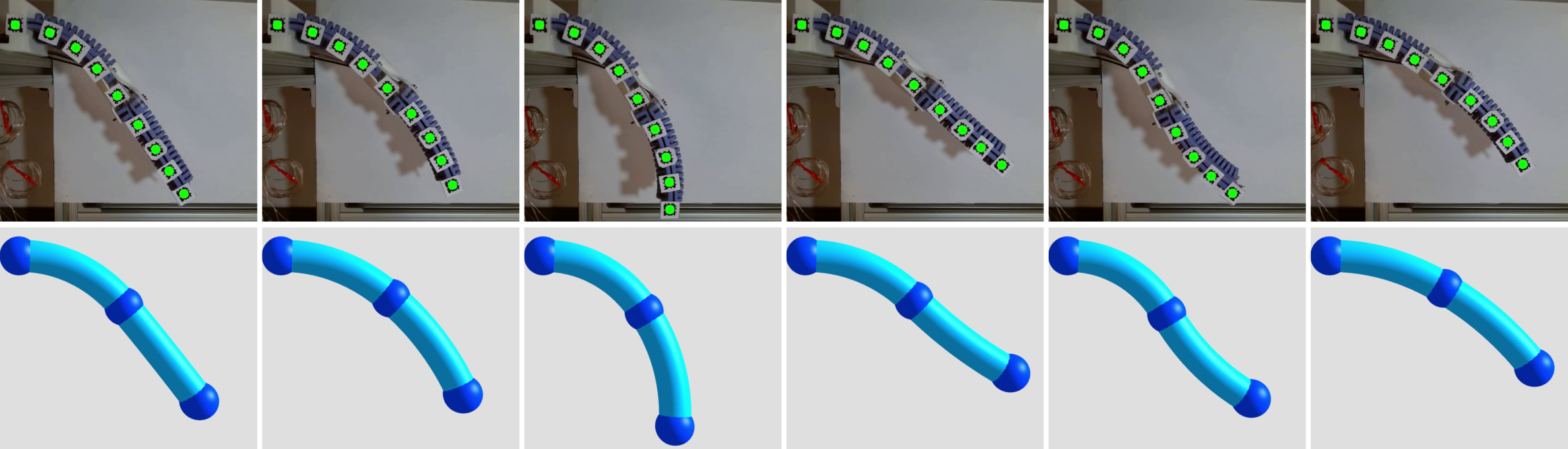}
	\caption{Snapshots showcasing real2sim realization of a SMA actuated dual soft limb manipulator~\cite{Pacheco2023comparison}. Using our gradient descent approach, we showcase excellent agreement between the real and simulated curvatures for a wide variety of geometric configurations.} 
\label{fig:real2sim}
\end{figure*}

\subsection{Real2Sim Open-Loop Control}
\label{sec:real2sim}

Finally, we provide a method to use DisMech to simulate an open-loop trajectory of a real soft robot in hardware, as a form of open-loop control.
Here, we use a shape memory alloy (SMA) actuated double limb soft manipulator~\cite{Pacheco2023comparison} (Fig.~\ref{fig:real2sim}).
The robot's limbs are $76 \times 49 \times 9$\,mm, and are constructed from a silicone polymer (Smooth-On Smooth-Sil) with a density of $1240$\,kg/m$^3$ and a Young's modulus of $1.793$\,MPa.
Despite the manipulator being quite wide, we can still represent the manipulator as a rod since manipulation occurs primarily in 2D~\cite{choi2024learning}.
For the rod radius, we used $10$\,cm to compensate for the ridges of the limbs. 
Two rods are initialized with the aforementioned parameters to represent each limb, which are then connected via an elastic joint.

\begin{figure}
    \centering
	\includegraphics[width=0.48\textwidth]{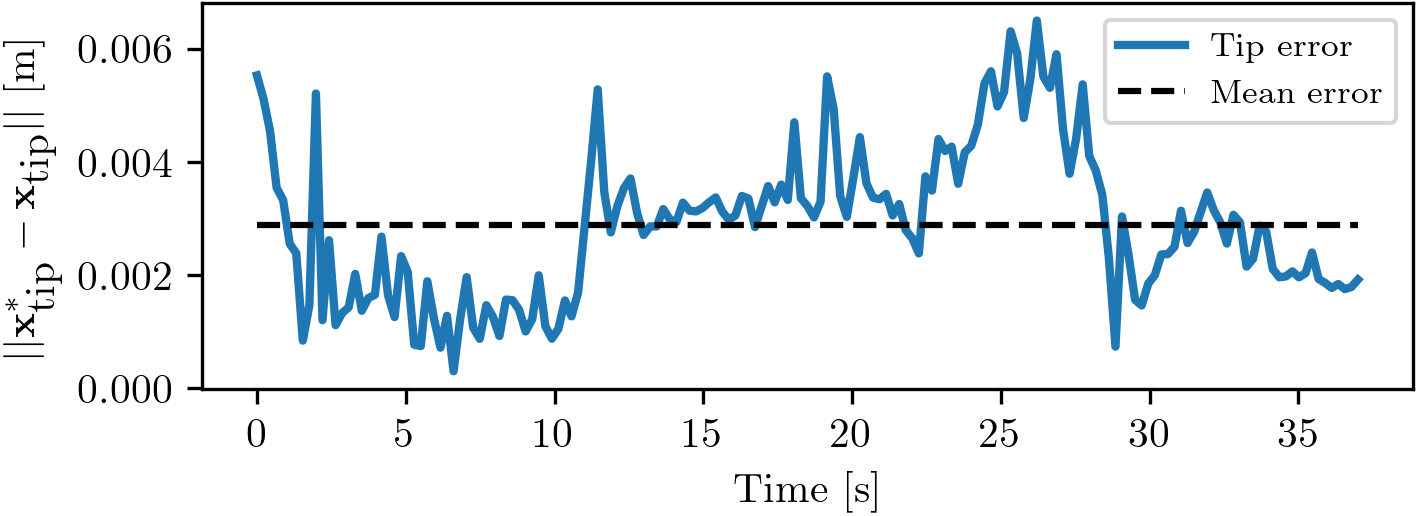}
	\caption{Tip error between the target tip position $\mathbf x^*_\textrm{tip}$ and the simulated tip position $\mathbf x_\textrm{tip}$ when carrying out our gradient descent natural curvature actuation.} 
\label{fig:real2sim_error}
\end{figure}

\subsubsection{Solving Natural Curvature via Gradient Descent}

To achieve real2sim realization, we must first calculate the appropriate $\bar {\boldsymbol \kappa}$ values that generate  geometric configurations of the robot corresponding to hardware data (Sec.~\ref{sec:natural_curvature_control}).
Given the nonlinearity of the robot's geometry, coupled with deformations produced by gravity, solving for the appropriate natural curvatures analytically is nontrivial. 
Therefore, we propose a gradient descent-based approach to solve for $\bar {\boldsymbol \kappa}$.  

In our approach, we assume that the movement of the limbs is quasistatic and that each limb has more-or-less a constant curvature along its length.
We use the AprilTag library as in our prior work \cite{Pacheco2023comparison} to extract the arm's position from video  (Fig.~\ref{fig:real2sim}, green dots).
Using these positions, we can then compute the target curvatures $\boldsymbol \kappa^*$ using Eq.~\ref{eq:compute_kappa}.

We then define a loss
\begin{equation}
\label{eq:lambda}
    \lambda(\bar \kappa_1, \bar \kappa_2) = | \kappa_1^* - \kappa_1 | + | \kappa_2^* - \kappa_2 |,
\end{equation}
where $\kappa_1$ and $\kappa_2$ are the resulting simulated curvatures of the first and second limbs when changing the natural curvatures to values $\bar \kappa_1$ and $\bar \kappa_2$, respectively.
The gradient of $\lambda$ with respect to $\bar{\boldsymbol \kappa}$ can then be obtained using a forward finite difference approach: 
\begin{equation}
\label{eq:finite_diff}
    \nabla_{\bar{\boldsymbol \kappa}} \lambda = \begin{bmatrix}
        \dfrac{\partial \lambda}{\partial \bar \kappa_1} \\ \dfrac{\partial \lambda}{\partial \bar \kappa_2}
    \end{bmatrix}
    = \frac{1}{\epsilon}\begin{bmatrix} 
        \lambda(\bar \kappa_1 + \epsilon, \bar \kappa_2) - \lambda(\bar \kappa_1, \bar \kappa_2) \\
        \lambda(\bar \kappa_1 , \bar \kappa_2 + \epsilon) - \lambda(\bar \kappa_1, \bar \kappa_2)
    \end{bmatrix},
\end{equation}
where $\epsilon$ is a small input perturbation. 
Using this finite difference gradient, we can then iteratively solve for the correct $\bar{\boldsymbol \kappa}$ by updating
\begin{equation}
\label{eq:grad_descent_update}
\bar{\boldsymbol \kappa} = \bar{\boldsymbol \kappa} - \alpha \nabla_{\bar{\boldsymbol \kappa}} \lambda
\end{equation}
until $\lambda$ reaches below a set tolerance, where $\alpha$ is a step size.
The full psuedocode for this approach can be seen in Alg.~\ref{alg:real2sim}.

We demonstrate excellent real2sim realization of the soft limb actuator through a DisMech model both visually (Fig.~\ref{fig:real2sim}) and numerically (Fig.~\ref{fig:real2sim_error}), where we achieve an average tip position error of just $2.9$\,mm ($1.5\%$ of the robot's length).
Future work, using more data, will derive separate actuator models (e.g. SMA voltages to natural curvatures), which in turn will allow for feedback controller design via sim2real. 

\section{Conclusion}
\label{sec:conclusion}

In this work, we introduced DisMech, a fully generalizable, implicit, discrete differential geometry-based physical simulator capable of accurate and efficient simulation of soft robots and structures composed of elastic rods.
We physically validated DisMech through several simulations, using both representative examples and new and complicated robots, showcasing an order of magnitude computational gain over previous methods.
In addition, we also introduced a method for intuitively actuating soft continuum robots by bending energy manipulation via natural curvature changes, and demonstrated this actuation's use as a control input in real2sim of a soft manipulator.

Future work will involve integrating shells~\cite{grinspun_2003_discrete_shells} into DisMech to allow for more complex deformable assemblies.
In addition to this, the incorporation of shearing into DisMech will be a key research area to allow it to simulate all deformation modes similar to Cosserat rod-based frameworks~\cite{gazzola2018forward, mathew2022sorosim}.
In fact, shearing has previously been integrated into a DDG-based framework when simulating a Timoshenko beam~\cite{li_2020_timo}, albeit for just a 2D framework.
Finally, efficient pipelines for real2sim2real modelling and training through reinforcement learning using DisMech will be a key research area moving forward for soft robot control and deformable object manipulation.

\balance 
\bibliographystyle{ieeetr}
\bibliography{arXiv}

\end{document}